\documentclass[sigconf, nonacm]{acmart}

\AtBeginDocument{%
  }

\setcopyright{acmlicensed}
\copyrightyear{2026}
\acmYear{2026}
\acmDOI{XXXXXXX.XXXXXXX}

\usepackage{color}
\usepackage{graphicx}
\usepackage{amsthm}
\usepackage{amsmath}
\usepackage{amsfonts}
\usepackage[ruled,noend,linesnumbered]{algorithm2e}
\usepackage{algpseudocode}
\usepackage{textcomp}
\usepackage{booktabs}
\usepackage[disable]{todonotes}
\usepackage{hyperref}
\usepackage{multirow}
\usepackage{tabularx}
\usepackage{tabularray}
\usepackage{subcaption}
\usepackage{listings}
\usepackage{xcolor}

\theoremstyle{plain}
\theoremstyle{definition}

\theoremstyle{remark}

\begin{document}

\title{Retrieval-Augmented Generation with Covariate Time Series}

\author{Kenny Ye Liang}
\authornote{K. Liang and Z. Pei contributed equally in this research.}
\affiliation{%
    \institution{Tsinghua University}
    \city{}
    \country{}
}
\email{liangy24@mails.tsinghua.edu.cn}

\author{Zhongyi Pei}
\authornotemark[1]
\affiliation{%
    \institution{Tsinghua University}
    \city{}
    \country{}
}
\email{peizhyi@tsinghua.edu.cn}

\author{Huan Zhang}
\affiliation{%
    \institution{China Southern Airlines}
    \city{}
    \country{}
}
\email{zhanghuan_a@csair.com}

\author{Yuhui Liu}
\affiliation{%
    \institution{China Southern Airlines}
    \city{}
    \country{}
}
\email{liuyh@csair.com}

\author{Shaoxu Song}
\authornote{Shaoxu Song (\url{https://sxsong.github.io/}) is the corresponding author.}
\affiliation{%
    \institution{Tsinghua University}
    \city{}
    \country{}
}
\email{sxsong@tsinghua.edu.cn}

\author{Jianmin Wang}
\affiliation{%
	\institution{Tsinghua University}
	\city{}
	\country{}
}
\email{jimwang@tsinghua.edu.cn}

\begin{abstract}
While RAG has greatly enhanced LLMs, extending this paradigm to Time-Series Foundation Models (TSFMs) remains a challenge.
This is exemplified in the Predictive Maintenance of the Pressure Regulating and Shut-Off Valve (PRSOV), a high-stakes industrial scenario characterized by (1) data scarcity, (2) short transient sequences, and (3) covariate coupled dynamics.
Unfortunately, existing time-series RAG approaches predominantly rely on generated static vector embeddings and learnable context augmenters, which 
may fail to distinguish similar regimes in such scarce, transient, and covariate coupled scenarios.
To address these limitations, we propose \textbf{RAG4CTS}, a regime-aware, training-free \textbf{RAG} framework for \textbf{C}ovariate \textbf{T}ime-\textbf{S}eries.
Specifically, we construct a hierarchal time-series native knowledge base to enable lossless storage and physics-informed retrieval of raw historical regimes.
We design a two-stage bi-weighted retrieval mechanism that aligns historical trends through point-wise and multivariate similarities.
For context augmentation, we introduce an agent-driven strategy to dynamically optimize context in a self-supervised manner.
Extensive experiments on PRSOV demonstrate that our framework significantly outperforms state-of-the-art baselines in prediction accuracy.
The proposed system is deployed in Apache IoTDB within China Southern Airlines.
Since deployment, our method has successfully identified one PRSOV fault in two months with zero false alarm.

\end{abstract}


\maketitle

\section{Introduction}
\label{sect:introduction}

The success of Large Language Models (LLMs) \cite{DBLP:conf/nips/BrownMRSKDNSSAA20,DBLP:journals/corr/abs-2302-13971} has spurred the development of Time Series Foundation Models (TSFMs) \cite{DBLP:journals/tmlr/AnsariSTZMSSRPK24,DBLP:conf/icml/LiuZLH0L24,DBLP:journals/corr/abs-2401-13912}.
However, while promising in open domains, Time-Series Foundation Models (TSFMs) struggle in high-stakes industrial applications due to a fundamental capability gap. The heterogeneous and unfamiliar data distributions in these settings lie outside the pre-training experience of TSFMs, preventing them from adapting to specialized system dynamics. The result is a failure to maintain physical consistency, which renders their predictions unreliable for critical decision-making.

\begin{figure}
    \centering
    \begin{subfigure}[b]{0.48\linewidth}
        \centering
        \includegraphics[width=\linewidth]{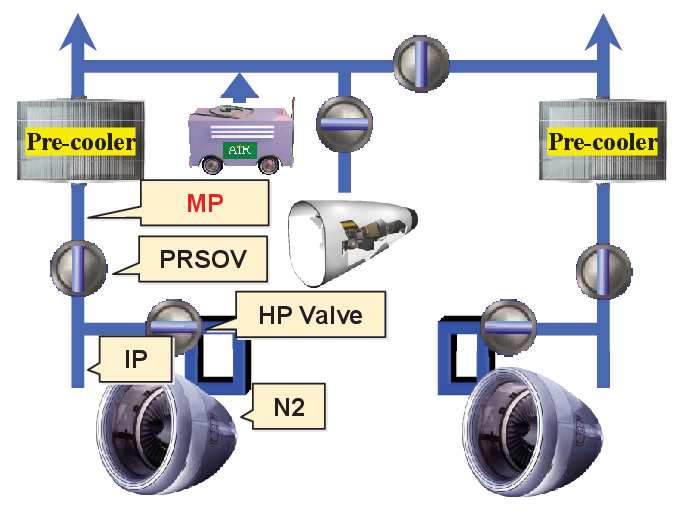}
        \caption{PRSOV Regime Mechanism.}
        \label{fig:motivation_phy}
    \end{subfigure}
    \hfill
    \begin{subfigure}[b]{0.48\linewidth}
        \centering
        \includegraphics[width=\linewidth]{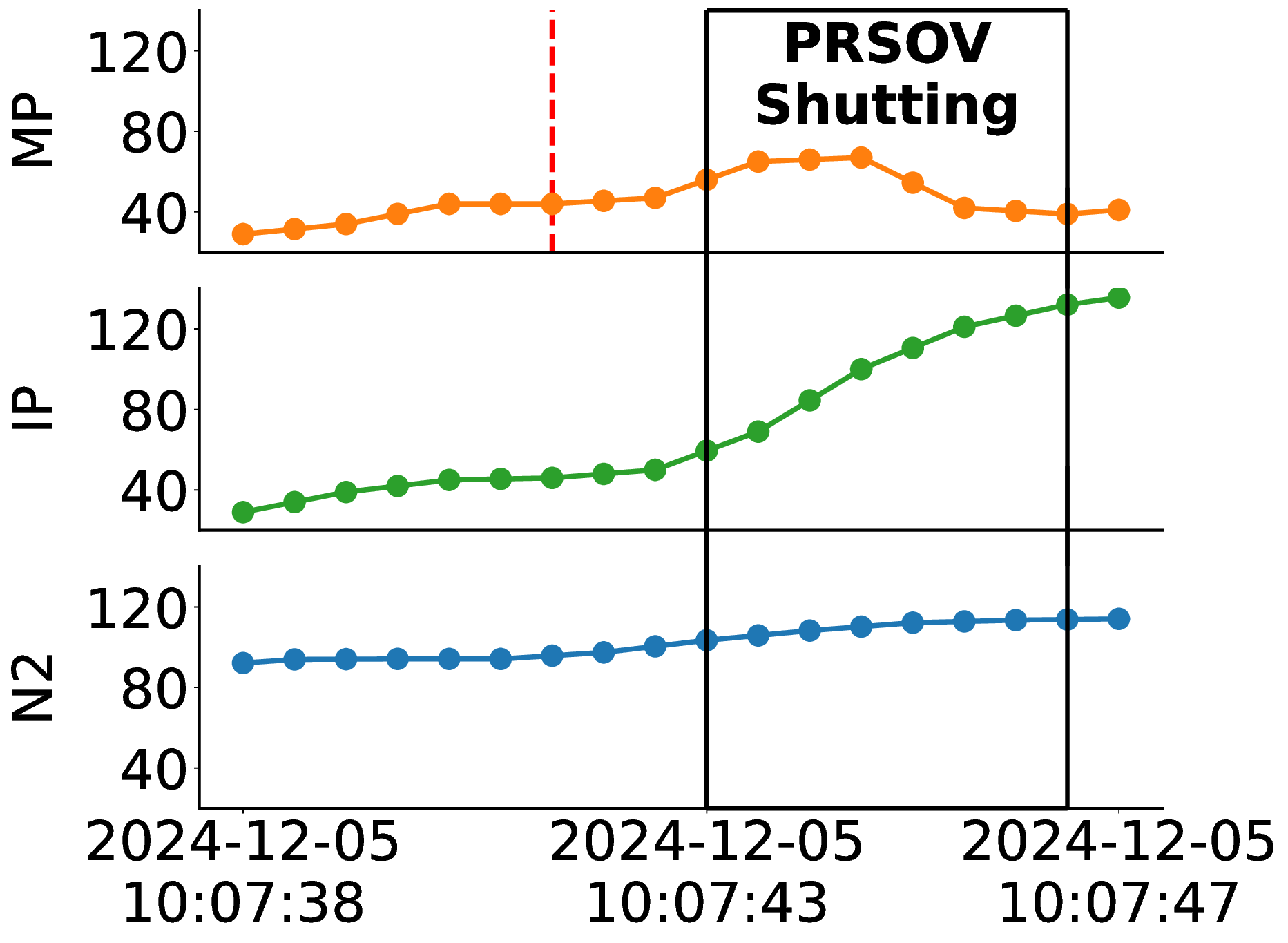} 
        \caption{PRSOV Regime Time-Series.}
        \label{fig:motivation_data}
    \end{subfigure}
    
    \caption{\textbf{Illustration of the PRSOV Scenario.} (a) The PRSOV operates under strict pneumatic control logic where the target Manifold Pressure (MP) is primarily influenced by the Engine Speed (N2) and Intermediate Pressure (IP). (b) Example of real world PRSOV: data scarcity (one sample per flight), short transient sequences (18 points in 10 seconds), and complex covariate coupling.}
    \label{fig:motivation_all}
\end{figure}

\begin{figure*}
    \centering
    \includegraphics[width=\linewidth]{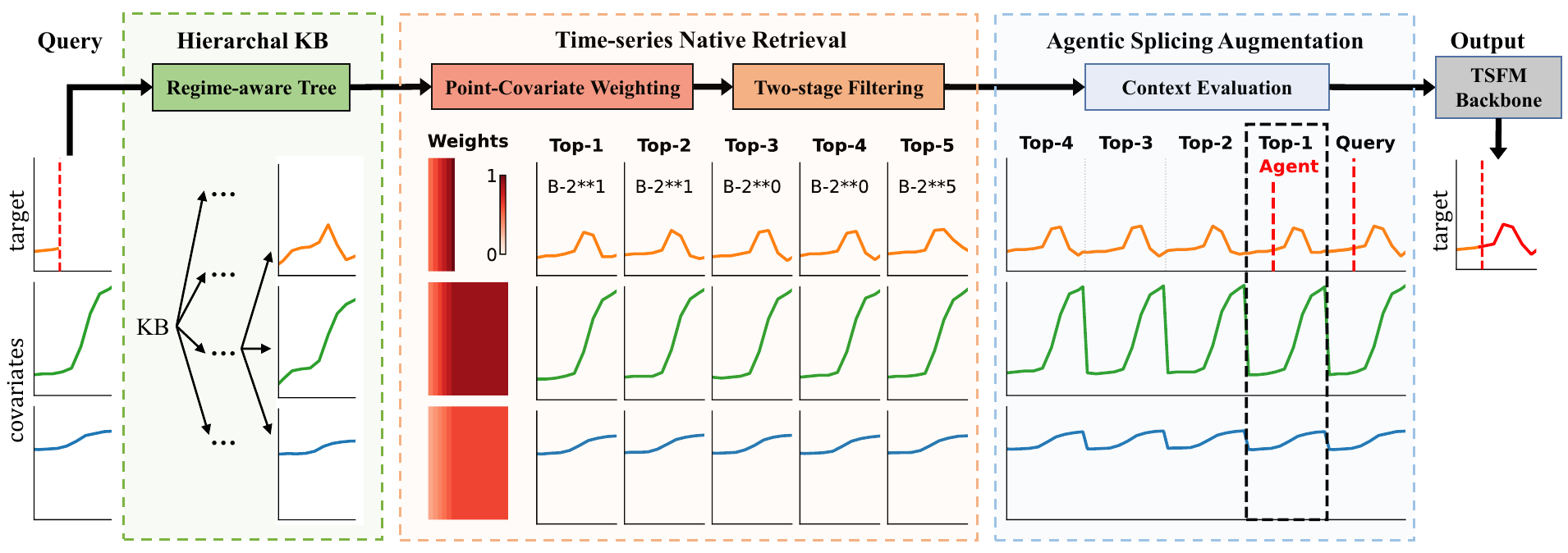}
    \caption{Overall RAG Pipeline. Tail numbers are masked for privacy.}
    \label{fig:method_overall_framework}
\end{figure*}

\subsection{Motivation}

This gap is particularly acute in the Predictive Maintenance of the \emph{Pressure Regulating and Shut-Off Valve} (PRSOV) in commercial aircraft.
Since internal valve degradation is unobservable, maintenance relies on monitoring deviations between the observed \emph{Manifold Pressure} (MP) and its estimated healthy baseline.
As illustrated in Figure \ref{fig:motivation_phy}, the system operates under a rigid chain where the regulated MP is passively driven by external covariates: the \emph{Engine High-Pressure Rotor Speed} (N2) and the upstream \emph{Intermediate Pressure} (IP).

Scenarios such as PRSOV pressure regulation reveal three defining characteristics that challenge conventional forecasting paradigms:
(1) \textbf{Data Scarcity.}
Critical operational regimes are inherently rare, creating a severe paucity of training data. This scarcity prevents deep models from learning meaningful representations. In the case of PRSOV, for example, its key regulation phase occurs only once per entire flight cycle.
(2) \textbf{Short Transient Context.}
The dynamics are characterized by rapid state transitions within extremely brief time windows. As illustrated in Figure \ref{fig:motivation_data}, a typical PRSOV regulation regime may comprise as few as 18 data points, a context too limited for most models to extract reliable temporal patterns.
(3) \textbf{Covariate Coupled Dynamics.}
The target variable (MP) is not autonomously determined but is passively driven by external covariates, namely N2 and IP. Attempting to forecast MP without explicitly modeling these governing relationships violates the system's inherent physical logic.

Inspired by Retrieval-Augmented Generation (RAG) in NLP 
\cite{DBLP:conf/nips/LewisPPPKGKLYR020}, we adapt the paradigm shift from parametric pattern matching to non-parametric in-context reference. This approach provides a direct solution for such challenging industrial time series forecasting:
(1) 
By leveraging retrieved, relevant historical sequences, the model learns from context without requiring data-hungry fine-tuning, thus directly overcoming the limitation of scarce regimes.
(2) 
The method augments brief information-sparse input windows with rich retrieved contexts from similar past events. This transforms an underdetermined forecasting task into a context-aware inference problem.
(3) 
It retrieves historical segments driven by identical covariate forces (i.e., N2 and IP), providing the model with physics-aligned references. 
These segments serve as contextual ``ground truth,'' ensuring predictions adhere to the system's inherent response logic.

\subsection{Challenge}
Despite the theoretical potential, existing RAG frameworks like TimeRAF \cite{DBLP:journals/tkde/ZhangXZZWB25} and TS-RAG \cite{DBLP:journals/corr/abs-2503-07649} suffer from fundamental architectural mismatches towards scenarios with scarce, short sequence, and covariate.
(1) Reliance on learnable adapters creates a data paradox. 
These data-hungry modules fail to generalize in scare settings, rendering the solution to scarcity dependent on the very data it lacks. 
For the selected PRSOV regime, which occurs only once per flight cycle, these trainable components struggle to converge and fail to capture the regime's unique characteristics.
(2) Static vectorization distorts short transients. 
To accommodate standard static embedding models, short transient sequences must be heavily padded, drowning the fine-grained signal in noise and obliterating the numerical precision required for transient analysis.
For PRSOV regimes containing only 18 points, padding to fixed-length inputs (e.g., 64 points) introduces dominant artifactual noise, effectively burying the critical transient signature needed for precise fault detection.
(3) Target-only retrieval ignores covariate logic. 
Visually similar target trajectories can be coincidental.
Without aligning driving forces, their future developments might inevitably diverge, rendering them invalid references.
For instance, MP fluctuations induced by High-Pressure Valve switching in other phases may visually mimic takeoff transients.
Blindly retrieving such contexts without aligning covariates N2 and IP can lead to referencing a completely unrelated operational regime.

\subsection{Contribution}
\label{subsect:contribution}
To overcome these limitations, we propose a novel framework designed for complex industrial covariate time-series.
To the best of our knowledge, this is the first study on TSFM-oriented RAG with covariate time-series.
Our contributions are summarized as follows:

(1) 
We construct a \textbf{time-series native knowledge base} using a tree-structured schema. We bypass the traditional vectorization of time-series, enabling lossless storage of raw operational regimes and preserving the full numerical precision of short transients for direct model ingestion.

(2) We design a \textbf{two-stage bi-weighted retrieval mechanism}. 
By exploiting known future covariates, we align historical segments through point-wise and multivariate similarities. 
This dual-weighting ensures retrieved contexts share the query's driving control intent, enforcing strict covariate logic.

(3) We introduce an \textbf{agent-driven robust context augmentation} strategy. 
Replacing static learnable context augmenters, our self-supervised method uses the Top-1 retrieval as a dynamic agent to calibrate the augmented context, achieving better performance.

(4) We present a fully deployed system on \textbf{Apache IoTDB} within \textbf{China Southern Airlines}. 
Extensive testing shows that our framework achieves the highest accuracy across three categories of time-series forecasting methods. 
Since deployment, it successfully identified a confirmed PRSOV fault in two months with zero false alarms, validating its industrial reliability.

\section{Related Work}
\label{sect:related_work}

\subsection{Deep Learning Approaches for Time Series}
Early deep learning methods primarily relied on Recurrent Neural Networks (RNNs) like LSTM \cite{DBLP:journals/neco/HochreiterS97} and Convolutional Neural Networks (CNNs) such as TCN \cite{DBLP:journals/corr/abs-2408-15737} to capture temporal dependencies. 
The introduction of Transformers shifted the focus towards attention mechanisms \cite{DBLP:conf/nips/VaswaniSPUJGKP17}. 
\textbf{Informer} \cite{DBLP:conf/aaai/ZhouZPZLXZ21} addressed the quadratic complexity of self-attention in long sequences.
\textbf{Autoformer} \cite{DBLP:conf/nips/WuXWL21} and \textbf{Pyraformer} \cite{DBLP:conf/iclr/LiuYLLLLD22} addressed computational complexity with decomposition and sparse attention.
Although \textbf{DLinear} \cite{DBLP:conf/aaai/ZengCZ023} challenged this trend with simple linear layers, Transformer variants regained SOTA status through \textbf{PatchTST} \cite{DBLP:conf/iclr/NieNSK23} (patching) and \textbf{iTransformer} \cite{DBLP:conf/iclr/LiuHZWWML24} (inverted dimensions).
Most recently, \textbf{TimeMixer} \cite{DBLP:conf/iclr/WangWSHLMZ024} and \textbf{TimeXer} \cite{DBLP:conf/nips/WangWDQZLQWL24} have introduced multi-scale mixing and external variable integration strategies.
These models serve as robust supervised baselines in our experiments, representing the state-of-the-art in fixed-schema learning.

\subsection{Time-Series Foundation Model (TSFM)}
TSFMs typically fall into two categories: LLM adaptations and native models.
Adaptation methods like \textbf{Time-LLM} \cite{DBLP:conf/iclr/0005WMCZSCLLPW24} and \textbf{LLMTime} \cite{DBLP:conf/nips/GruverFQW23} reprogram text-based LLMs for time series via prompting or token alignment.
Conversely, native LTSMs are pretrained from scratch on massive time-series corpora.
\textbf{Timer} \cite{DBLP:conf/icml/LiuZLH0L24} and \textbf{TimerXL} \cite{DBLP:conf/iclr/LiuQ00L25} adopt a decoder-only architecture for scalable generative forecasting.
The \textbf{Chronos} family \cite{DBLP:journals/tmlr/AnsariSTZMSSRPK24}, including the recent \textbf{Chronos-Bolt} and \textbf{Chronos-2} \cite{DBLP:journals/corr/abs-2510-15821}, quantizes continuous values into discrete tokens to leverage cross-entropy objectives similar to language modeling.
Other notable native models include \textbf{Moirai} \cite{DBLP:conf/icml/WooLKXSS24} for universal forecasting and \textbf{Sundial} \cite{DBLP:conf/icml/LiuQSCY00L25} for its temporal attention.
Despite their zero-shot potential, these models often suffer from hallucinations in specific industrial contexts, highlighting the need for retrieval augmentation.

\subsection{Time-Series RAG}
Retrieval-Augmented Generation (RAG) aims to mitigate the context limitations of TSFMs.
Initial attempts like \textbf{TimeRAG} \cite{DBLP:conf/icassp/YangWZJ25} utilize Dynamic Time Warping (DTW) to retrieve similar raw sequences as text prompts for LLMs.
More complex frameworks include \textbf{TimeRAF} \cite{DBLP:journals/tkde/ZhangXZZWB25}, which employs a ``Retriever-Forecaster Joint Training'' paradigm to align historical embeddings, and \textbf{TS-RAG} \cite{DBLP:journals/corr/abs-2503-07649}, which introduces an Adaptive Retrieval Mixer (ARM) to fuse retrieved semantic embeddings with the model's internal representations.
In contrast, our approach fundamentally diverges from these embedding-centric methods by operating directly in the raw data space in a fully zero-shot, training-free manner.

\section{Methodology}
\label{sect:methodology}

To address the core objective of high-reliability time series analysis under complex regime-switching conditions, we propose RAG4CTS, a regime-aware native RAG framework for Covariate Time-Series, as shown in Figure \ref{fig:method_overall_framework}.
The framework consists of three key components: 
(1) A Hierarchical Knowledge Base for lossless raw regime storage, displayed in Section \ref{subsect:method_kb}. 
(2) A time-series native two-stage bi-weighted retrieval mechanism, established in Section \ref{subsect:method_retriever}.
(3) An agentic splicing strategy for self-supervised context augmentation, detailed in Section \ref{subsect:method_aug}.

\begin{figure}
    \centering
    \includegraphics[width=0.9\linewidth]{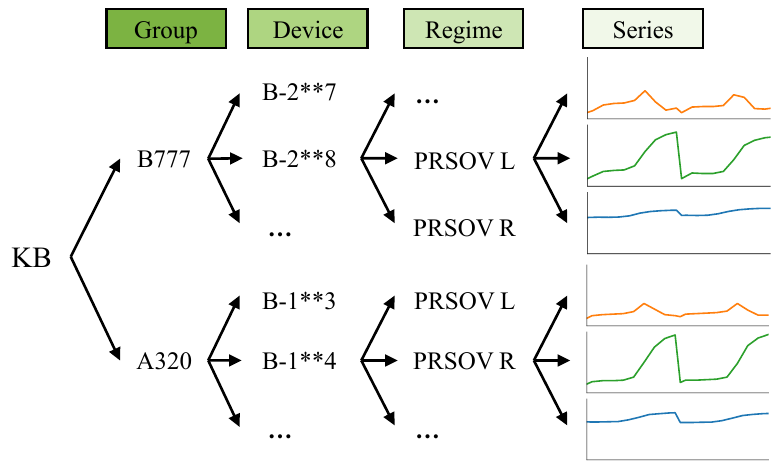}
    \caption{Tree-Structured Knowledge Base. Unlike vector stores, it preserves raw sequences following their physical hierarchy. Tail numbers are masked for privacy.}
    \label{fig:method_kb}
\end{figure}

\subsection{
Hierarchical
Knowledge Base}
\label{subsect:method_kb}

To facilitate precise retrieval without information loss, we construct a regime-aware hierarchical knowledge base $\mathcal{B}$.
Formally, let $\mathbf{X} \in \mathbb{R}^{T \times V}$ denote a multivariate time series with $T$ timestamps and $V$ variables.
We define the knowledge base as a set of $N$ historical samples of the operational regime:
\begin{equation}
    \mathcal{B} = \{ (\mathbf{M}_i, \mathbf{X}_i) \}_{i=1}^N
\end{equation}
where $\mathbf{M}_i$ is the hierarchical path (e.g., root.group.device.sensor), acting as a metadata index. 
$\mathbf{X}_i$ represents the complete raw recording of a specific operational cycle (e.g., a full PRSOV regime in the aircraft takeoff phase).

Figure \ref{fig:method_kb} illustrates the architecture tailored for PRSOV.
The structure descends from group identifiers (e.g., aircraft type B777) to device identifiers (e.g., aircraft tail numbers) to specific regime instances (e.g., PRSOV L), storing raw multivariate series at the leaf nodes.
Unlike vector databases requiring slicing or padding, our native storage preserves the full integrity of each operational cycle. 
This design ensures that retrieved candidates retain their mechanical independence and physical integrity, and are free of the artificial fragmentation and approximation errors inherent in embedding-based approaches.

\subsection{Time-Series Native Retrieval Mechanism}
\label{subsect:method_retriever}

Existing RAG approaches slice and map time series into static vector embeddings. 
This global compression inherently obscures absolute magnitudes and treats all points uniformly, thereby failing to capture the point-wise correlations and covariate coupling that are essential for physical systems.
To address this, we propose a two-stage bi-weighted retrieval mechanism operating directly in the raw data space.
This mechanism acts as a physical prior, decomposing retrieval alignment into two dimensions:
(1) \textbf{Critical Point Weighting}, which prioritizes recent system states and known future controls, and
(2) \textbf{Covariate Weighting}, which emphasizes driving variables with strong causal influence.
By integrating these weights, we ensure retrieved contexts are not merely visually similar but share the same critical control logic as the query.

\subsubsection{Critical Point Weighting}
In transient analysis, normally, not all timestamps carry equal information density.
The most recent system states and the immediate future control inputs serve as Critical Points that determine the trajectory's evolution.
We construct a critical point weight matrix $\mathbf{W}^{\text{point}} \in \mathbb{R}^{L \times V}$, where $L$ is the total length of the sequence window (history + future horizon), to explicitly model this significance.
Let $\mathcal{V}_{\text{cov}}$ denote the driving covariates with known future values, and $\mathcal{V}_{\text{target}}$ denote the target variable.
The weight logic aligns retrieval with future intentions while strictly masking the unknown targets.
Crucially, since the query lacks future target values (the prediction goal), this zero-masking ensures the distance is computed solely on the shared available information (history and future covariates), ignoring the ground truth present in the KB candidates:
\begin{equation}
    \mathbf{W}^{\text{point}}_{t, v} = 
    \begin{cases} 
      \lambda^{(L_{\text{hist}}-t)} & \text{if } t \le L_{\text{hist}} \quad \text{(History Decay)} \\
      1 & \text{if } t > L_{\text{hist}} \land v \in \mathcal{V}_{\text{cov}} \quad \text{(Future Control)} \\
      0 & \text{if } t > L_{\text{hist}} \land v \in \mathcal{V}_{\text{target}} \quad \text{(Target Masking)}
   \end{cases}
\end{equation}
where $\lambda \in (0, 1]$ is a decay factor prioritizing recent dynamic patterns.

\subsubsection{Covariate Weighting}
In coupled physical systems, covariates exert uneven causal influence on the target.
Even among active covariates, their impact varies by physical proximity. 
For instance, while both Intermediate Pressure (IP) and Rotor Speed (N2) influence Manifold Pressure (MP), IP exerts a stronger, immediate pneumatic impact due to its structural closeness, whereas N2 acts as a secondary upstream covariate.
To quantify this hierarchy, we employ Mutual Information (MI) \cite{DBLP:journals/tnn/Battiti94} to construct the covariate weight vector $\mathbf{w}^{\text{cov}}$.
Unlike linear correlations, MI captures non-linear physical couplings by measuring the reduction in uncertainty of the target $Y$ given a covariate $X_v$ in a training-free manner.

Formally, let $X_v$ ($v \in \mathcal{V}_{\text{cov}}$) be the sequence of a specific covariate and $Y$ be the target. Their Mutual Information is defined as:
\begin{equation}
    I(X_v; Y) = \sum_{x \in X_v} \sum_{y \in Y} p(x,y) \log \left( \frac{p(x,y)}{p(x)p(y)} \right)
\end{equation}
We compute this score globally across the knowledge base $\mathcal{B}$ to capture robust statistical dependencies. 
The final covariate weight vector $\mathbf{w}^{\text{cov}} \in \mathbb{R}^{V}$ is then normalized to prioritize the most informative sensors:
\begin{equation}
    \mathbf{w}^{\text{cov}}_v =
    \begin{cases}
        1.0 & \text{if } v \in \mathcal{V}_{\text{target}} \\
        \frac{I(X_v; Y)}{\max_{k \in \mathcal{V}_{\text{cov}}} I(X_k; Y)} & \text{if } v \in \mathcal{V}_{\text{cov}}
    \end{cases}
\end{equation}
This normalization ensures that the primary physical covariate (e.g., IP) serves as the dominant reference anchor alongside the target variable itself, effectively filtering out noise from weakly coupled sensors.
This also assigns equal importance to the target variable during retrieval, preventing the driving logic from being overshadowed by the target history.

\subsubsection{Unified Bi-Weighted Fusion}
To synthesize these two dimensions, the final retrieval weight matrix $\mathbf{W} \in \mathbb{R}^{L \times V}$ is computed via the Hadamard product ($\odot$) of the critical point matrix and the broadcast covariate vector:
\begin{equation}
    \mathbf{W} = \mathbf{W}^{\text{point}} \odot \mathbf{w}^{\text{cov}} \quad \text{i.e., } \mathbf{W}_{t, v} = \mathbf{W}^{\text{point}}_{t, v} \cdot \mathbf{w}^{\text{cov}}_v
\end{equation}
This fusion minimizes retrieval distance specifically at structurally critical time steps and physically dominant variables.

\begin{figure}%
    \centering
    \includegraphics[width=\linewidth]{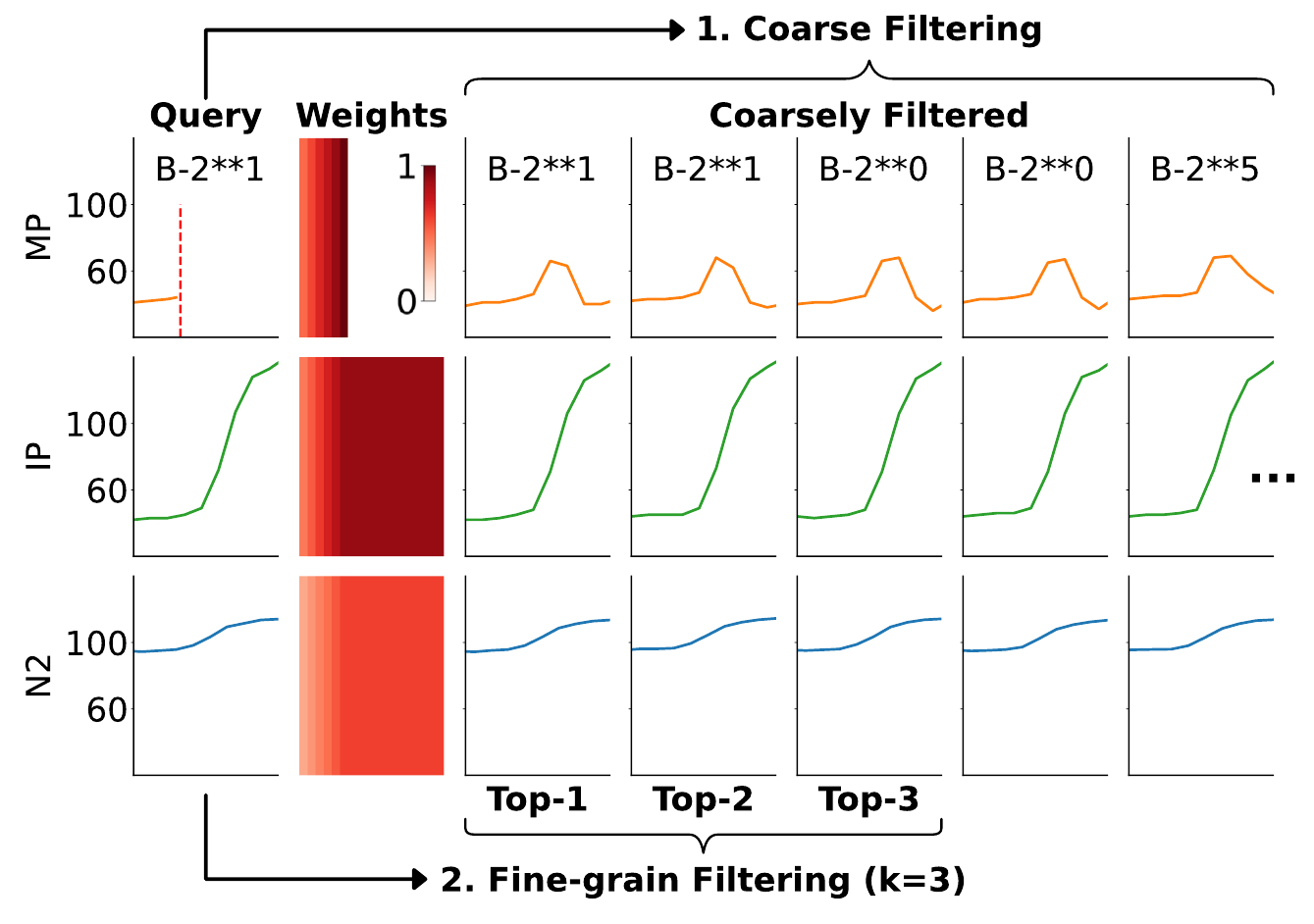}
    \caption{The Time-Series Native Retrieval Mechanism. It employs a bi-weighted coarse-to-fine strategy. Tail numbers are masked for privacy.}
    \label{fig:hybrid_retrieval}
\end{figure}

\subsubsection{Two-stage Retrieval Filtering}

Relying on a single similarity metric often fails to capture the multifaceted nature of time-series fidelity. 
Correlation metrics capture shape but ignore magnitude, while distance metrics are sensitive to absolute states but may overlook structural alignment.
To ensure comprehensive physical consistency, we employ a two-stage filtering strategy \cite{DBLP:conf/fodo/AgrawalFS93} that progressively enforces both qualitative trend alignment and quantitative state precision.

\textbf{Stage 1: Shape Alignment via Weighted Correlation.}
The first stage focuses on identifying candidates that share the same evolutionary trend (e.g., specific transient edges) regardless of baseline shifts.
We employ the Cosine Similarity \cite{DBLP:journals/cacm/SaltonWY75} between the query $\mathbf{Q}$ and each candidate $\mathbf{C}_i \in \mathcal{B}$ using our bi-weighted matrix $\mathbf{W}$:
\begin{equation}
    \mathcal{S}_{\text{shape}}(\mathbf{Q}, \mathbf{C}_i; \mathbf{W}) = \frac{\sum_{t,v} \mathbf{W}_{t,v} \cdot \mathbf{Q}_{t,v} \cdot \mathbf{C}_{i,t,v}}{ \sqrt{\sum_{t,v} \mathbf{W}_{t,v} (\mathbf{Q}_{t,v})^2} \cdot \sqrt{\sum_{t,v} \mathbf{W}_{t,v} (\mathbf{C}_{i,t,v})^2} }
\end{equation}
We select the Top-$10K$ candidates with the highest $\mathcal{S}_{\text{shape}}$ scores into an intermediate set $\mathcal{C}_{\text{shape}}$. 
This step acts as a morphological filter, ensuring that retrieved contexts are structurally isomorphic to the query's control logic.

\textbf{Stage 2: State Precision via Weighted Matrix Profile.}
From the shape-aligned candidates $\mathcal{C}_{\text{shape}}$, we aim to pinpoint the exact matches that minimize physical state discrepancy.
We adopt the \textbf{Matrix Profile} distance metric \cite{DBLP:conf/icdm/YehZUBDDSMK16}, enhanced with our unified weighting scheme, to measure absolute numerical fidelity.
The weighted distance $D_{\text{mp}}$ between the query $\mathbf{Q}$ and a candidate $\mathbf{C}_i \in \mathcal{C}_{\text{shape}}$ is defined as:
\begin{equation}
    D_{\text{mp}}(\mathbf{Q}, \mathbf{C}_i; \mathbf{W}) = \sqrt{ \sum_{t=1}^{L} \sum_{v=1}^{V} \mathbf{W}_{t,v} \cdot \left( \mathbf{Q}_{t,v} - \mathbf{C}_{i, t,v} \right)^2 }
\end{equation}
The final output of the retrieval phase is formally structured as an ordered set of the Top-$K$ candidates with the smallest $D_{\text{mp}}$:
\begin{equation}
    \mathcal{C}_{\text{final}} = \{ \mathbf{X}_{r_1}, \dots, \mathbf{X}_{r_K} \} \quad \text{s.t.} \quad D_{\text{mp}}^{(r_1)} \le D_{\text{mp}}^{(r_2)} \le \dots \le D_{\text{mp}}^{(r_K)}
\end{equation}
By integrating trend-based correlation and magnitude-sensitive Matrix Profile distance under a unified weighting scheme, our mechanism ensures high-fidelity references that are causally aligned with the future control logic.

Figure \ref{fig:hybrid_retrieval} illustrates the execution of this mechanism.
The heatmaps visualize our bi-weighted logic, assigning maximal importance (dark red) to the target's recent trajectory and the covariates' future control inputs.
Following the two-stage filtering strategy, the system first retrieves a broad set of shape-aligned candidates, from which the weighted distance metric pinpoints the final Top-3 ($K=3$) references that share identical physical driving forces.

\begin{figure}
    \centering
    \includegraphics[width=\linewidth]{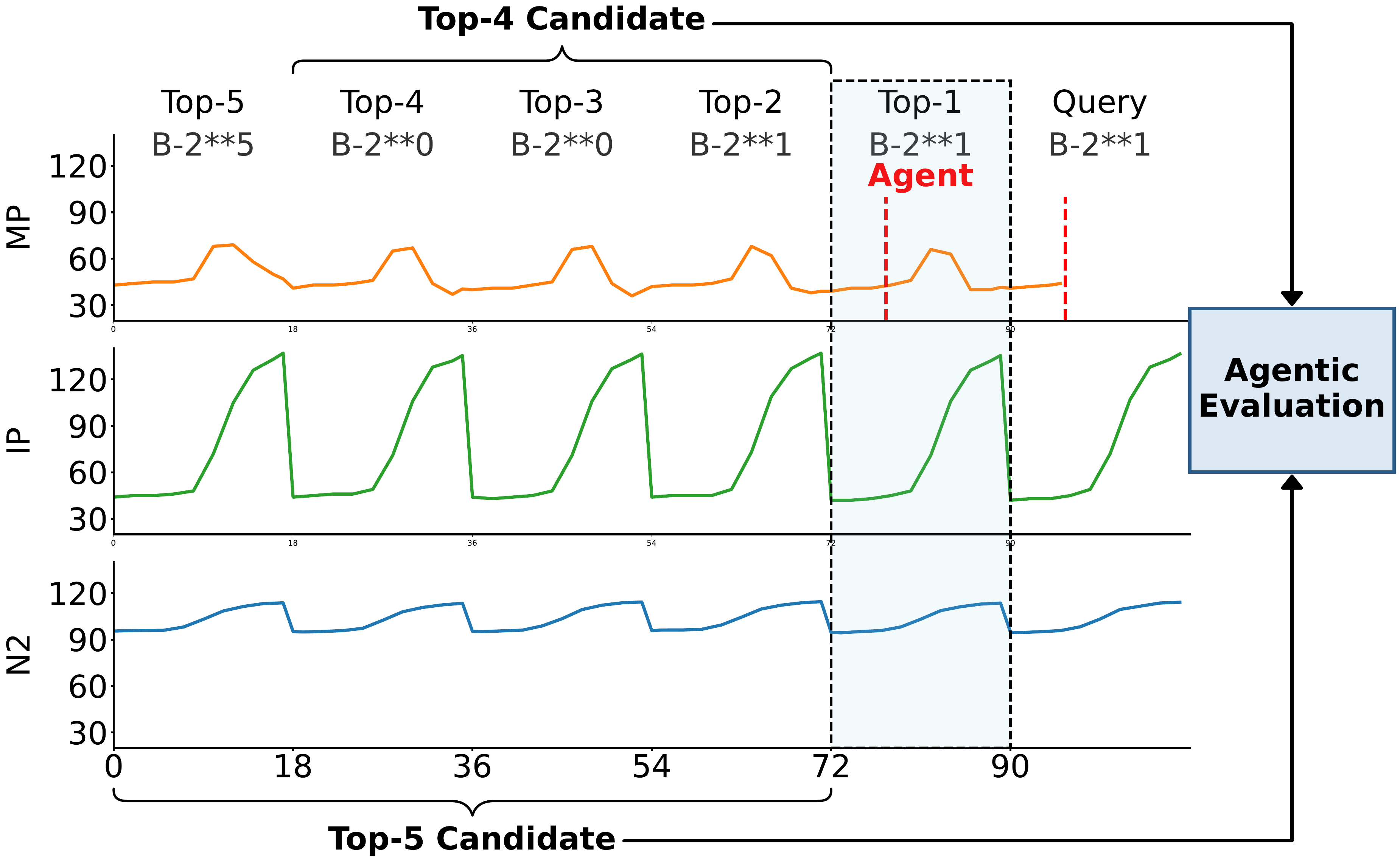}
    \caption{The agentic context augment process. The top-1 sample is used as an agent to self-calibrate the optimal number of context fragments ($k^*$). Tail numbers are masked for privacy.}
    \label{fig:agent_fusion}
\end{figure}

\subsection{Agent-driven Context Augmentation}
\label{subsect:method_aug}

To effectively utilize the retrieved contexts $\mathcal{C}_\text{final} = \{\mathbf{X}_{r_1}, \dots, \mathbf{X}_{r_K}\}$, we leverage the In-Context Learning (ICL) capabilities of foundation models \cite{DBLP:conf/nips/BrownMRSKDNSSAA20, DBLP:journals/corr/abs-2510-15821}, which adapt via prompt demonstrations without gradient updates.
Accordingly, we employ raw value splicing, directly concatenating retrieved regimes before the query as physical demonstrations.
However, context quantity does not equate to quality: insufficient context fails to activate ICL, whereas excessive context introduces attention noise.
Simply maximizing context length is suboptimal and instance-dependent.
Existing methods typically rely on a fixed hyperparameter $K$ or learned adapters, which are either suboptimal or data-inefficient.

To address this, we propose an agentic context-optimization strategy. 
We treat the inference process as a self-reflective loop.
We designate the most relevant retrieved sample, the Top-1 regime sample $\mathbf{X}_{r_1} \in \mathcal{C}_\text{final}$, as an Agent Query, utilizing its known future $\mathbf{Y}_{r_1}$ as the ground truth for calibration. 
The remaining samples $\{\mathbf{X}_{r_2}, \dots, \mathbf{X}_{r_K}\}$ serve as the candidate context pool.

The agent performs a greedy search to determine the optimal context count $k^*$ that minimizes prediction error.
It progressively prepends retrieved segments to the agent query in reverse rank order (i.e., placing the most relevant $\mathbf{X}_{r_2}$ closest to the agent $\mathbf{X}_{r_1}$), evaluating the loss at each step:
\begin{equation}
    k^* = \operatorname*{argmin}_{k \in \{1, \dots, K-1\}} \mathcal{L} \left( \mathcal{F}(\text{Concat}(\mathbf{X}_{r_{k+1}}, \dots, \mathbf{X}_{r_2}, \mathbf{X}_{r_1})), \mathbf{Y}_{r_1} \right)
\end{equation}
where $\mathcal{F}$ is the frozen TSFM and $\mathcal{L}$ is the prediction loss.

This specific splicing order $\mathbf{X}_{r_{k+1}} \oplus \dots \oplus \mathbf{X}_{r_2} \oplus \mathbf{X}_{r_1}$ ensures that the most relevant physical priors are positioned adjacent to the query, maximizing the attention mechanism's efficacy.
Once $k^*$ is calibrated on the agent query, the optimal context configuration is applied to the user query $\mathbf{X}_q$:
\begin{equation}
    \hat{\mathbf{Y}} = \mathcal{F}(\text{Concat}(\mathbf{X}_{r_1}, \dots, \mathbf{X}_{r_{k^*}}, \mathbf{X}_q))
\end{equation}

Figure \ref{fig:agent_fusion} visualizes this greedy iterative calibration using the Top-1 retrieved sample (shaded in light blue) as an agent.
It depicts the evaluation step in which the system tests the incremental benefit of splicing the Top-5 candidate into the existing context chain (Top-4 to Top-1).
By comparing the prediction error on the Agent's known future with that of the previous configuration, the framework dynamically determines the optimal cutoff point ($k^*$) at which information gain is maximized before noise degrades performance.

\begin{table}%
  \centering
  \caption{Statistics of the CSA-PRSOV Datasets. \textbf{KB} denotes Knowledge Base (2023-2024). \textbf{Min/Max} denotes the minimum and maximum sample counts per plane in KB.}
  \label{tab:dataset_stat}
  
  \resizebox{\columnwidth}{!}{
  
  \begin{tabular}{lccccccc}
    \toprule
    \multirow{2}{*}{\textbf{Dataset}} & \multirow{2}{*}{\textbf{Planes}} & \multirow{2}{*}{\textbf{Dim}} & \multirow{2}{*}{\textbf{Len}} & \multicolumn{3}{c}{\textbf{KB Samples}} & \textbf{Query} \\
    \cmidrule(lr){5-7} \cmidrule(lr){8-8}
     & & & & \textbf{Total} & \textbf{Min} & \textbf{Max} & \textbf{Total} \\
    \midrule
    \textbf{B777L} & 30 & 3 & 18 & 17,573 &  91 & 1,050 & 7,380 \\
    \textbf{B777R} & 30 & 3 & 18 & 17,573 &  91 & 1,050 & 7,380 \\
    \textbf{A320L} & 41 & 3 & 18 & 112,932 & 1,707  & 3,244 & 30,013 \\
    \textbf{A320R} & 41 & 3 & 18 & 112,932 & 1,707  & 3,244 & 30,013 \\
    \bottomrule
  \end{tabular}
  }
\end{table}

\begin{table*}
  \centering

  \newcolumntype{Y}{>{\centering\arraybackslash}X} 
  
  \caption{Main performance comparison. The best results are highlighted in \textbf{bold} and the second best are \underline{underlined}.}
  \label{tab:main_results}

  \begin{tabularx}{0.95\linewidth}{c l *{8}{Y}} 
        \toprule
        \multicolumn{2}{c}{\multirow{2}{*}{\textbf{Methods}}} & \multicolumn{2}{c}{\textbf{B777L}} & \multicolumn{2}{c}{\textbf{B777R}} & \multicolumn{2}{c}{\textbf{A320L}} & \multicolumn{2}{c}{\textbf{A320R}} \\
        
        \cmidrule(lr){3-4} \cmidrule(lr){5-6} \cmidrule(lr){7-8} \cmidrule(lr){9-10}
        
        \multicolumn{2}{c}{} & MSE & MAE & MSE & MAE & MSE & MAE & MSE & MAE \\
        \midrule
        
        \multirow{6}{*}{\rotatebox{90}{\textbf{Learning-based}}}
          & DLinear & 0.254 & 0.336 & 0.430 & 0.467 & 0.562 & 0.560 & 0.600 & 0.578 \\
          & TimeMixer & 0.467 & 0.452 & 0.580 & 0.538 & 0.509 & 0.512 & 0.570 & 0.538 \\
          & Pyraformer & \underline{0.085} & \underline{0.176} & \underline{0.108} & \textbf{0.168} & \underline{0.376} & \underline{0.427} & \underline{0.429} & \underline{0.457} \\ 
          & iTransformer & 0.171 & 0.252 & 0.181 & 0.270 & 0.471 & 0.491 & 0.554 & 0.530 \\
          & PatchTST   & 0.486 & 0.461 & 0.580 & 0.533 & 0.564 & 0.518 & 0.594 & 0.538 \\ 
          & TimeXer    & 0.262 & 0.316 & 0.225 & 0.310 & 0.452 & 0.478 & 0.512 & 0.509 \\ 
        \midrule
        
        \multirow{5}{*}{\rotatebox{90}{\textbf{TSFM}}} 
          & Sundial (zeroshot)  & 2.563 & 1.264 & 2.688 & 1.278 & 2.953 & 1.468 & 2.922 & 1.451 \\  
          & TimesFM (zeroshot)  & 2.346 & 1.180 & 2.428 & 1.187 & 2.854 & 1.429 & 2.835 & 1.416 \\ 
          & Chronos-2 (zeroshot)  & 1.542 & 0.907 & 1.586 & 0.913 & 2.013 & 1.155 & 1.984 & 1.103 \\
          & Chronos-Bolt-Base\textsubscript{B} (zeroshot) & 2.346 & 1.180 & 2.428 & 1.187 & 2.854 & 1.429 & 2.835 & 1.146 \\ 
          & Chronos-2 (finetuned)  & 0.296 & 0.286 & 0.463 & 0.416 & 1.234 & 0.602 & 1.268 & 0.690 \\
        \midrule

        \multirow{3}{*}{\rotatebox{90}{\textbf{RAG}}}  
          & TS-RAG\textsubscript{zeroshot} & 2.000 & 1.076 & 1.820 & 1.028 & 1.915 & 1.121 & 1.768 & 1.054 \\ 
          & TS-RAG\textsubscript{train}    & 0.960 & 0.711 & 1.776 & 0.937 & 1.611 & 1.033 & 1.521 & 0.995 \\ 
          & \textbf{RAG4CTS (ours)}  & \textbf{0.058} & \textbf{0.153} & \textbf{0.095} & \underline{0.203} & \textbf{0.259} & \textbf{0.331} & \textbf{0.337} & \textbf{0.384} \\
          
  \bottomrule
  \end{tabularx}%
\end{table*}

\begin{table}
  \centering
  \caption{Impact of covariate availability on forecasting performance. Full physical covariates yield the highest accuracy.}
  \label{tab:covariate_ablation}
  \resizebox{0.8\linewidth}{!}{%
    \begin{tabular}{l|cc|cc}
      \toprule
      \multirow{2}{*}{\textbf{Configuration}} & \multicolumn{2}{c|}{\textbf{RAG4CTS (ours)}} & \multicolumn{2}{c}{\textbf{Chronos-2}} \\
      \cmidrule(lr){2-3} \cmidrule(lr){4-5}
       & \textbf{MSE} & \textbf{MAE} & \textbf{MSE} & \textbf{MAE} \\
      \midrule
      0 covariate & 0.254 & 0.315 & 2.289 & 1.189 \\
      Only N2 & 0.249 & 0.319 & 1.958 & 0.972 \\
      Only IP & 0.214 & 0.296 & 1.729 & 1.020 \\
      \textbf{Full covariates} & \textbf{0.187} & \textbf{0.268} & \textbf{1.781} & \textbf{1.020} \\
      \bottomrule
    \end{tabular}%
  }
\end{table}

\begin{table}
  \centering
  \caption{Impact of Knowledge Base scope on B777. ``B777'' indicates the same aircraft type as the query.}
  \label{tab:kb_full}
  \resizebox{0.75\columnwidth}{!}{
    \begin{tabular}{lcccc}
      \toprule
      \multirow{2}{*}{\textbf{KB Strategy}} & \multicolumn{2}{c}{\textbf{B777L}} & \multicolumn{2}{c}{\textbf{B777R}} \\
      \cmidrule(lr){2-3} \cmidrule(lr){4-5}
       & \textbf{MSE} & \textbf{MAE}  & \textbf{MSE} & \textbf{MAE} \\
      \midrule
      Same Plane  & 0.103 & 0.205  & 0.216 & 0.314 \\
      B777 & 0.058 & \textbf{0.153} & \textbf{0.095} & \textbf{0.203} \\
      Full KB & \textbf{0.057} & 0.154 & 0.096 & 0.217 \\
      \bottomrule
    \end{tabular}%
  }
\end{table}

\section{Experiment}

We evaluate our framework on real-world PRSOV scenarios characterized by scarcity, short transients, and strong coupling.
Following the experimental setup in Section \ref{subsect:exp_setup}, we present the Overall Performance comparison against SOTA baselines in Section \ref{subsubsect:baseline_comparison}, followed by an in-depth Covariate Analysis in Section \ref{subsubsect:covariate_evaluation}.
We then conduct a comprehensive ablation study of the component design, covering the Knowledge Base (Section \ref{subsubsect:kb_evaluation}), Retrieval Mechanism (Section \ref{subsubsect:retrieve_evaluation}), and Context Augmentation (Section \ref{subsubsect:context_evaluation}).
Reproducibility details and visualizations are provided in Appendix \ref{appendix:reproducibility}.

\subsection{Experimental Setups}
\label{subsect:exp_setup}
We evaluate our framework on the CSA-PRSOV dataset from China Southern Airlines, divided into four subsets by aircraft type (B777/A320) and engine position (L/R).
The knowledge base is built with PRSOV samples from 2023 to 2024.
The samples from 2025 are used as queries.
As shown in Table \ref{tab:dataset_stat}, newer aircraft have only 91 historical samples.
The task involves short-term transient forecasting ($L=12,H=6$) of MP with two covariates (N2, IP).
We compare our method against three categories of baselines: (1) SOTA Deep Forecasters \cite{wang2024tssurvey}, 
e.g., PatchTST \cite{DBLP:conf/iclr/NieNSK23}, TimeMixer \cite{DBLP:conf/iclr/WangWSHLMZ024}; 
(2) Zero-shot TSFMs,
e.g., Chronos-2 \cite{DBLP:journals/corr/abs-2510-15821}; and 
(3) TS-RAG \cite{DBLP:journals/corr/abs-2503-07649}.
Our framework utilizes Chronos-2 as the TSFM backbone.
More details are provided in the Reproducibility Section (Appendix \ref{appendix:exp_setup}).

\subsection{Overall Comparison}
\label{subsect:overall_comparison}
In this subsection, we first benchmark our approach against SOTA methods.
Subsequently, we evaluate the influences of the covariates.

\subsubsection{Baseline Comparison}
\label{subsubsect:baseline_comparison}
Table \ref{tab:main_results} summarizes the quantitative results. 
Among all three categories of methods, our method consistently achieves the lowest errors, outperforming the second-best methods by a significant margin.

(1) Standard deep learning models generally struggle. 
Complex architectures like iTransformer fail to generalize on the small-scale RPSOV dataset.
Notably, \textbf{Pyraformer} performs best among them, as its sparse attention mechanism effectively captures local motifs in short sequences ($L+H=18$), unlike other method with global attention such as PatchTST.

(2) Time Series Foundation Models (TSFMs) like Chronos-2 (Zero-shot) perform poorly. 
Without a sufficient context window, TSFMs cannot infer the underlying periodicity or trend from just a few historical points. 
While fine-tuning improves performance, it still lags behind our method, indicating that weight adaptation alone cannot compensate for the information loss in short transient inputs.

(3) TS-RAG performs sub-optimally. 
Its reliance on learning a mapping from context to prediction fails in this regime because both the sequence length and the number of training samples are too small to train a robust adapter. 
It struggles to effectively utilize the retrieved information.

Our method outperforms Chronos-2 (Zero-shot) dramatically by turning a ``short-sequence'' problem into a ``long-context'' one.

\subsubsection{Covariate Evaluation}
\label{subsubsect:covariate_evaluation}
To validate the impact of the covariates (IP and N2) of MP, we compare different input combinations in Table \ref{tab:covariate_ablation}.
The results indicate that Chronos-2 exhibits a strong inherent capability to interpret covariates.
Adding them in the zero-shot setting notably reduces error, confirming that these physical sensors act as critical control signals.
Similarly, our RAG framework achieves peak performance with Full Covariates.
By incorporating our covariate weighting scheme, the retrieval prioritizes the dominant driving factors (e.g., IP).
This ensures the augmented context aligns with the target's primary control logic, enabling the model to predict PRSOV behavior with high fidelity.

\subsection{Ablation Study}
\label{subsect:ablation_study}
In this section, we individually evaluate the contribution of each module within our framework. 
We evaluate the impact of knowledge base scope, the efficacy of our physics-aware retrieval metrics, and the benefits of our dynamic context augmentation strategy.

\subsubsection{Knowledge Base Evaluation}
\label{subsubsect:kb_evaluation}

Table \ref{tab:kb_full} reveals a crucial trade-off. 
Expanding the KB from a single plane to the full fleet (``B777'') significantly reduces MSE ($0.103 \rightarrow 0.058$), validating that operational logic is transferable within the same aircraft type.
However, incorporating cross-type data (``B777 + A320'') yields diminishing returns.
Distinct airframe physics introduce distribution shifts that outweigh the benefits of increased data volume, suggesting that a model-specific KB is the optimal boundary for industrial retrieval.

\begin{table}
  \centering
  \caption{Sensitivity of retrieval metrics. The hybrid Cosine + Matrix Profile strategy is the most effective.}
  \label{tab:metric_ablation}
  
  \resizebox{0.72\linewidth}{!}{
    \begin{tabular}{lccc}
      \toprule
      \textbf{Retrieval Metric} & \textbf{MSE} & \textbf{MAE} &  \\
      \midrule
      Cosine + DTW       & 0.103 & 0.203 \\
      Cosine + Euclidean & 0.085 & 0.186 \\
      \textbf{Cosine + Matrix Profile}       & \textbf{0.077} & \textbf{0.178} \\
      Euclidean + DTW  & 0.111 & 0.208 \\
      Euclidean + Matrix Profile & \underline{0.080} & \underline{0.181} \\
      Matrix Profile + DTW & 0.099 & 0.197 \\
      \midrule 
      Cosine       & 0.082 & 0.185 \\
      Euclidean    & 0.085 & 0.187 \\ 
      Matrix Profile           & \underline{0.080} & \underline{0.181} \\ 
      \bottomrule
    \end{tabular}%
  }
\end{table}

\begin{table}
  \centering
  \caption{Ablation on weighting schemes and context length ($k$) on B777. ``$W^\text{point}$'' for Point Weighting and ``$W^\text{cov}$'' for Covariate Weighting.}
  \label{tab:ablation_k}
  
  \resizebox{\columnwidth}{!}{
    \begin{tabular}{lcccccc}
      \toprule
      \multirow{2}{*}{\textbf{Method Setup}} & \multicolumn{2}{c}{\textbf{Uniform}} & \multicolumn{2}{c}{\textbf{$W^\text{point}$}} & \multicolumn{2}{c}{\textbf{$W^\text{point} \odot W^\text{cov}$}} \\
      
      \cmidrule(lr){2-3} \cmidrule(lr){4-5} \cmidrule(lr){6-7}
      
       & MSE & MAE & MSE & MAE & MSE & MAE \\
      \midrule
      
      $k=0$ (Chronos-2) & 1.564 & 0.910 & 1.564 & 0.910 & 1.564 & 0.910 \\
      $k=1$  & 1.992 & 1.145 & 2.052 & 1.107 & 2.132 & 1.158 \\
      $k=2$  & 0.459 & 0.501 & 0.431 & 0.458 & 0.433 & 0.455 \\
      $k=3$  & 0.151 & 0.306 & 0.144 & 0.294 & 0.141 & 0.289 \\
      $k=6$  & 0.095 & 0.220 & 0.086 & 0.213 & 0.084 & 0.185 \\
      $k=9$  & 0.092 & 0.201 & 0.082 & 0.199 & 0.080 & 0.182 \\
      $k=12$ & 0.094 & 0.202 & 0.085 & 0.196 & 0.081 & 0.187 \\ 
      
      \textbf{Dynamic $k$} & 0.091 & 0.195 & 0.082 & 0.191 & \textbf{0.077} & \textbf{0.178} \\
      \bottomrule
    \end{tabular}%
  }
\end{table}

\subsubsection{Retrieval Mechanism Evaluation}
\label{subsubsect:retrieve_evaluation}

Table \ref{tab:metric_ablation} validates our hybrid retrieval strategy. 
Standard metrics alone are insufficient. 
Among them, Matrix Profile (MP), designed for motif discovery, is superior on its own. 
Overall, the hybrid ``Cosine + Matrix Profile'' combination achieves the lowest MSE (0.077) by balancing trend direction with precise shape matching.
Notably, adding DTW \cite{DBLP:conf/kdd/BerndtC94} degrades performance, as temporal warping distorts the rigid timing required for transient analysis.
This confirms our metric selection in the two-stage design.
Furthermore, Table \ref{tab:ablation_k} confirms the superiority of our Bi-Weighting strategy ($W^{\text{point}} \odot W^{\text{cov}}$), outperforming no weights at all (Uniform) and only point-wise weighting. 
By explicitly encoding the physical influence of covariates, the retriever identifies contexts that dynamically govern the target, rather than merely being visually similar.

\begin{figure}
    \centering
    \includegraphics[width=0.9\linewidth]{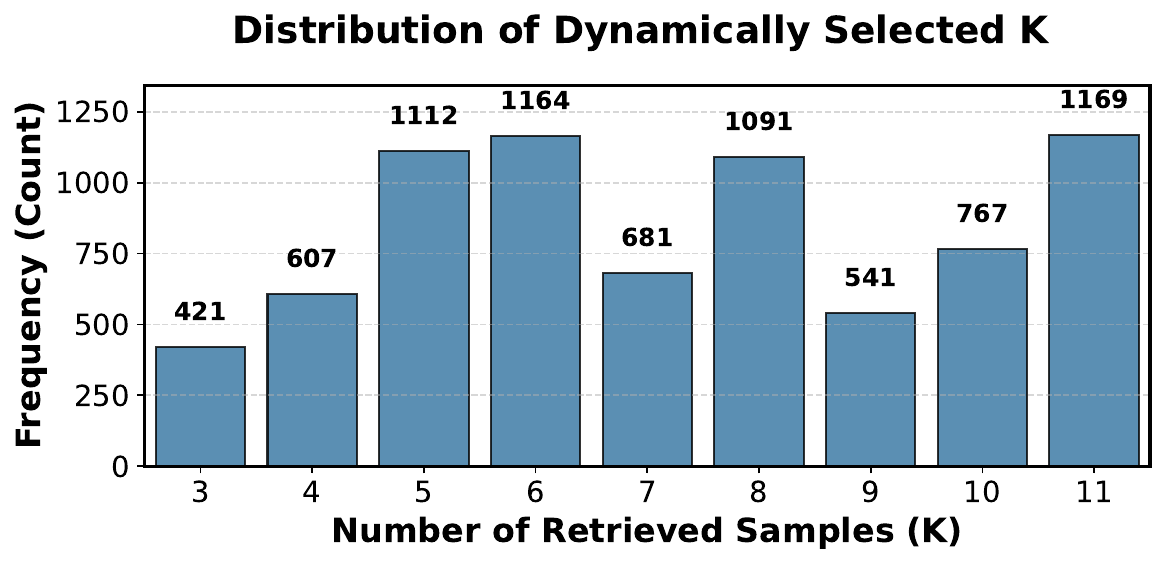}
    \caption{The distribution of k selected dynamically by our agentic splicing context augment.}
    \label{fig:k_distribution}
\end{figure}

\subsubsection{Context Augmentation Evaluation}
\label{subsubsect:context_evaluation}

Finally, we analyze the impact of context augmentation $k$ in Table \ref{tab:ablation_k}.
Extremely short contexts ($k=1$) often introduce noise without sufficient pattern repetition, while increasing $k$ generally improves accuracy by providing more historical cycles.
Our agentic context augmentation strategy achieves the lowest error (MSE 0.077).
Instead of a fixed number of retrieval contexts, it dynamically selects augmentation to maximize prediction stability. 
Figure \ref{fig:k_distribution} illustrates the high variance in dynamically selected $k$ values across queries. 
This confirms that a fixed retrieval context number is suboptimal and highlights our method's ability to adaptively tailor the retrieval depth to the specific complexity of each query.

\section{Deployment}

We have successfully integrated our RAG framework into the operational maintenance workflow of China Southern Airlines.
Section \ref{subsect:deploy_iotdb} details the architecture of the system deployed on the AINode of Apache IoTDB.
Section \ref{subsect:deploy_csa_case} then presents the deployment outcomes.

\subsection{Apache IoTDB and AINode}
\label{subsect:deploy_iotdb}

Our framework is deployed in the \textbf{AINode} \cite{apache_iotdb_ainode}, the native machine learning engine of \textbf{Apache IoTDB} \cite{DBLP:journals/pacmmod/0018QHSHJR0023}, enabling in-database model inference via standard SQL to minimize data migration overhead.
Crucially, we integrated the proposed RAG framework to support declarative industrial forecasting.
As shown below, engineers can use our RAG4CTS for Manifold Pressure (MP) by specifying the driving covariates (N2, IP) in an SQL query.
The implementation code is available in the open-source repository \cite{deployed_iotdb}.

\begin{verbatim}
IoTDB > SELECT * from
    forecast(
        model_id => 'RAG4CTS',
        input_table => 'B777',
        devices_filter => device_id = 'tail#',
        target => 'MP',
        covariates => 'IP' AND 'N2',
        output_start_time => 2026-02-01,
        output_interval => 0.5s,
        output_length => 12);
\end{verbatim}

\begin{figure}
    \centering
    \includegraphics[width=\linewidth]{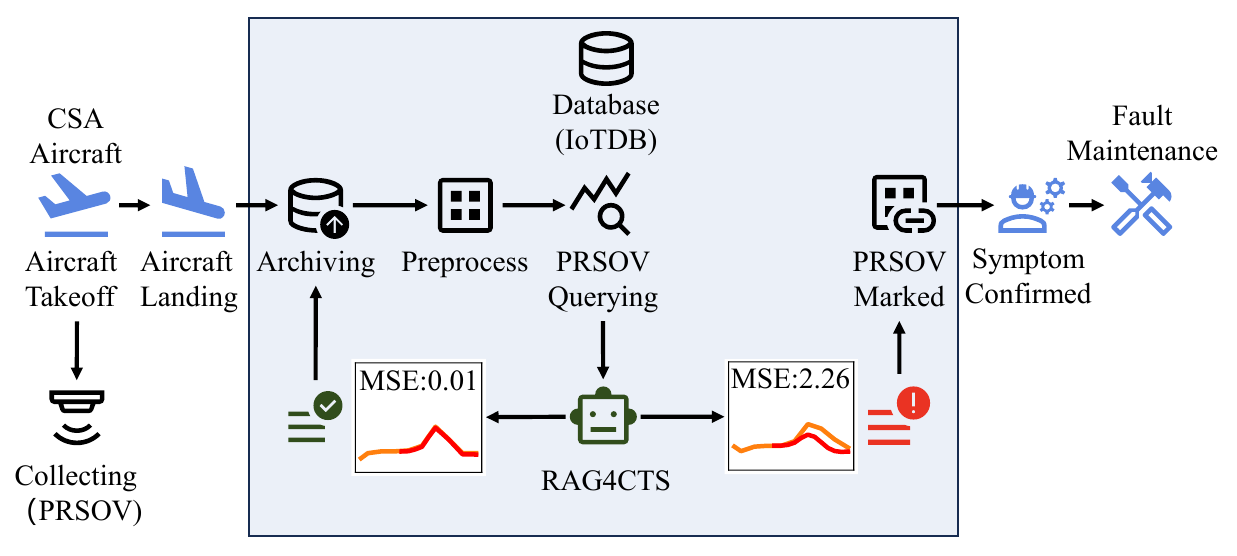}
    \caption{The PRSOV predictive maintenance pipeline of China Southern Airlines.}
    \label{fig:case_study_pipeline}
\end{figure}

\begin{table}%
  \centering
  \caption{Predictive maintenance records from the CSA deployment (Starting Nov. 2025). Online visualizations are presented in Appendix \ref{appendix:case_study_illustration}. Tail numbers are masked for privacy.}
  \label{tab:case_study_records}
  \resizebox{\columnwidth}{!}{
    \begin{tabular}{lcccc}
      \toprule
      \textbf{Tail} & \textbf{Fault Type} & \textbf{Identified Date} & \textbf{Actual Fault} & \textbf{Type}\\
      \midrule
      B-2**8 & PRSOV R & 2023-10-15 & Yes & Historical \\
      B-2**0 & PRSOV R & 2023-11-04 & Yes & Historical \\
      B-7**5 & PRSOV R & 2023-12-26 & Yes & Historical \\
      B-2**1 & PRSOV L & 2024-04-01 & Yes & Historical \\
      B-2**0 & PRSOV R & 2024-08-22 & Yes & Historical \\
      B-2**9 & PRSOV R & 2024-09-11 & Yes & Historical \\
      B-2**1 & PRSOV L & 2024-10-2  & Yes & Historical \\
      B-2**Y & PRSOV R & 2024-10-26 & Yes & Historical \\
      B-2**G & PRSOV R & 2024-12-28 & Yes & Historical \\
      B-2**N & PRSOV R & 2025-02-22 & Yes & Historical \\
      \midrule
      B-2**7 & PRSOV L & 2025-12-31  & Yes (Fig.~\ref{fig:case_study_b2027}) & Online \\
      \bottomrule
    \end{tabular}%
  }
\end{table}

\subsection{China Southern Airlines Online Case Study}
\label{subsect:deploy_csa_case}

To address the limitations of reactive maintenance, we integrated our framework into the aircraft health management platform at China Southern Airlines (CSA). 
The system has been deployed since late November 2025.

\subsubsection{Operational Challenge}
At China Southern Airlines (CSA), the maintenance strategy for the PRSOV system has historically been reactive. 
Prior to our work, reliance was placed on internal self-checks conducted immediately prior to takeoff. 
If a fault is detected at this critical juncture, it will inevitably lead to Aircraft on Ground (AOG) events or technical delays until emergency repairs are performed. 
According to Boeing's operational metrics, a single technical delay of this nature incurs an estimated financial loss of \$50,000, including damage to brand reputation. 
In response to these problems, CSA sought to transition to a \textbf{Proactive Predictive Maintenance} strategy. 
The objective is to utilize forecasting models to identify potential fault precursors days in advance, allowing maintenance to be scheduled during routine layovers to ensure aircraft safety and operational continuity.
Figure \ref{fig:case_study_pipeline} illustrates this end-to-end workflow designed for PRSOV maintenance.
After aircraft landed, the data collected is archived into Apache IoTDB, where raw time-series data undergo automated preprocessing to isolate relevant transient regimes.
Subsequently, our RAG4CTS framework is triggered to forecast the expected behavior for the queried PRSOV segment.
Based on the forecasting results (MSE), the system either archives the data as a normal sample or tags high-deviation instances for engineering verification.

\subsubsection{Forecasting and Alerting}
Adapting the academic model for the industrial environment required two specific engineering configurations:
(1) In production, the Retriever is restricted to a curated Knowledge Base of healthy historical regimes. 
Consequently, the RAG model functions as a predictor of ``ideal behavior''.
If the model fails to reconstruct the current activation (resulting in a high generation error) using only healthy references, it indicates a physical deviation from the norm.
(2) To eliminate false alarms caused by sensor noise, single anomalies rarely trigger immediate alerts in CSA. 
Instead, the system monitors a two-week rolling window. 
A ``Fault Precursor'' is confirmed only when the frequency of high-generation-error deviations significantly exceeds the baseline. 
This logic is critical for distinguishing intermittent degradation signals from transient operational noise.

\subsubsection{Operational Results}
The deployment outcomes are summarized in Table \ref{tab:case_study_records}. 
Upon deployment, the system was rigorously backtested on historical flights (2023–2025) and successfully identified 10 fault precursors, all of which were subsequently verified against the following fault recordings.
Since its online deployment in late November 2025, the system has monitored the fleet in real-time.
Through December 2025, it flagged one aircraft, B-2**7, as a potential risk for PRSOV fault.
This alert was subsequently confirmed by engineering teams as a genuine PRSOV fault, and no false alarms were triggered.
Detailed visualization of the precursor pattern for the online case is provided in Appendix \ref{appendix:case_study_illustration}.

\section{Conclusion}
This paper addresses the critical limitations of current TSFMs and RAG methods in handling scarce, covariate-coupled industrial time series, specifically for PRSOV predictive maintenance.
To bridge this gap, we introduce \textbf{RAG4CTS}, a regime-aware native \textbf{RAG} framework for \textbf{C}ovariate \textbf{T}ime-\textbf{S}eries, featuring a hierarchical knowledge base, a two-stage bi-weighted retrieval mechanism, and an agentic augmentation strategy.
Extensive experiments confirm that our approach significantly outperforms state-of-the-art baselines in forecasting accuracy.
Furthermore, its deployment at \textbf{China Southern Airlines} has proven its industrial value: successfully identified a PRSOV fault with zero false alarms over two months.
These results validate the framework as a robust, scalable solution for transitioning complex industrial systems from reactive to proactive maintenance.

\begin{acks}
This work is supported in part by 
the National Key Research and Development Plan (2025ZD1601701, 2024YFB3311901, 2021YFB3300500), 
the National Natural Science Foundation of China (62232005, 92267203, 62021002), 
Beijing National Research Center for Information Science and Technology (BNR2025RC01011), and
Beijing Key Laboratory of Industrial Big Data System and Application. 
Shaoxu Song (\url{https://sxsong.github.io/}) is the corresponding author.
\end{acks}

\bibliographystyle{plain}
\bibliography{bib/references}

\appendix

\section{Reproducibility}
\label{appendix:reproducibility}
In this section, we provide the detailed experimental settings and the qualitative visualizations to complement the quantitative results presented in the main experiments. 
Specifically, the detailed experimental setups is in Appendix \ref{appendix:exp_setup}
Then, we examine the behavior of our framework under different configurations, directly corresponding to the subsections in the main text: Appendix \ref{appendix:rag_vis} Baseline Comparison (Section \ref{subsubsect:baseline_comparison}), Appendix \ref{appendix:covariate_vis} Covariate Evaluation (Section \ref{subsubsect:covariate_evaluation}), Appendix \ref{appendix:kb_vis}
Knowledge Base Scope (Section \ref{subsubsect:kb_evaluation}), Appendix \ref{appendix:weighting_vis}
Retrieval Mechanism (Section \ref{subsubsect:retrieve_evaluation}), and 
Appendix \ref{appendix:topk_vis}
Context Augmentation (Section \ref{subsubsect:context_evaluation}). 
These visualizations illustrate how specific components contribute to the prediction accuracy in CSA-PRSOV scenarios.
The MSE values reported in the top-right corner represent the normalized MSE. 
The parameter K indicates the actual number of retrieved segments concatenated with the query. 
Note that for visual clarity, only the top-5 retrieved samples are displayed.

\subsection{Experimental Setups}
\label{appendix:exp_setup}

\subsubsection{Datasets and Knowledge Base Construction}
We evaluate our framework on the CSA-PRSOV dataset, a real-world industrial collection from the Pressure Regulating and Shut-Off Valve systems of China Southern Airlines.
To rigorously test generalization across different physical configurations, we construct four distinct sub-datasets based on aircraft model (Boeing 777 vs. Airbus A320) and engine position (Left vs. Right).

The dataset is constructed with a strict temporal split: historical flight cycles from 2023 to 2024 constitute the Knowledge Base (KB), while flights from 2025 onward constitute the Query set.
This setup introduces a realistic ``Cold Start'' challenge for newer aircraft. 
As shown in Table \ref{tab:dataset_stat}, while the mature B777 aircraft have over 1,000 historical cycles, newer entries have as few as \textbf{91 samples}. 
This 10x data imbalance makes it difficult for global models to capture the specific degradation patterns of newer engines, demonstrating the scarce characteristic for these tail-end assets.
The forecasting task targets the transient takeoff phase with a rapid 10 second engine spool-up window, resulting in extremely short sequences of only 18 data points per cycle.
The objective is to predict the bleed Manifold Pressure (MP) for the next 12 steps given the preceding 6 steps ($L=6, H=12$).

\subsubsection{Baselines}
We compare our framework against three categories of methods:
(1) \textbf{Deep Learning-based Models}: SOTA supervised forecasters including PatchTST \cite{DBLP:conf/iclr/NieNSK23}, iTransformer \cite{DBLP:conf/iclr/LiuHZWWML24}, TimeMixer \cite{DBLP:conf/iclr/WangWSHLMZ024}, TimeXer \cite{DBLP:conf/nips/WangWDQZLQWL24}, Pyraformer \cite{DBLP:conf/iclr/LiuYLLLLD22}, and DLinear \cite{DBLP:conf/aaai/ZengCZ023}, trained from scratch
(2) \textbf{Zero-shot TSFMs}: Zero-shot Large Time Series Models including Chronos-2 \cite{DBLP:journals/corr/abs-2510-15821, hf_chronos_2}, Chronos-Bolt \cite{hf_chronos_bolt}, and Sundial \cite{DBLP:conf/icml/LiuQSCY00L25}.
(3) \textbf{Time-Series RAG}: The vector-embedding based TS-RAG \cite{DBLP:journals/corr/abs-2503-07649}.

\subsubsection{Implementation Details}
Our native RAG framework is deployed online on \textbf{Apache IoTDB}. 
We use Chronos-2 without finetuning as the backbone TSFM.
Models' parameters followed the same settings as Time-Series Library \cite{wang2024tssurvey}.
The results can be reproduced with the same settings \cite{deployed_iotdb}.

\begin{figure}
    \centering
    \includegraphics[width=\linewidth]{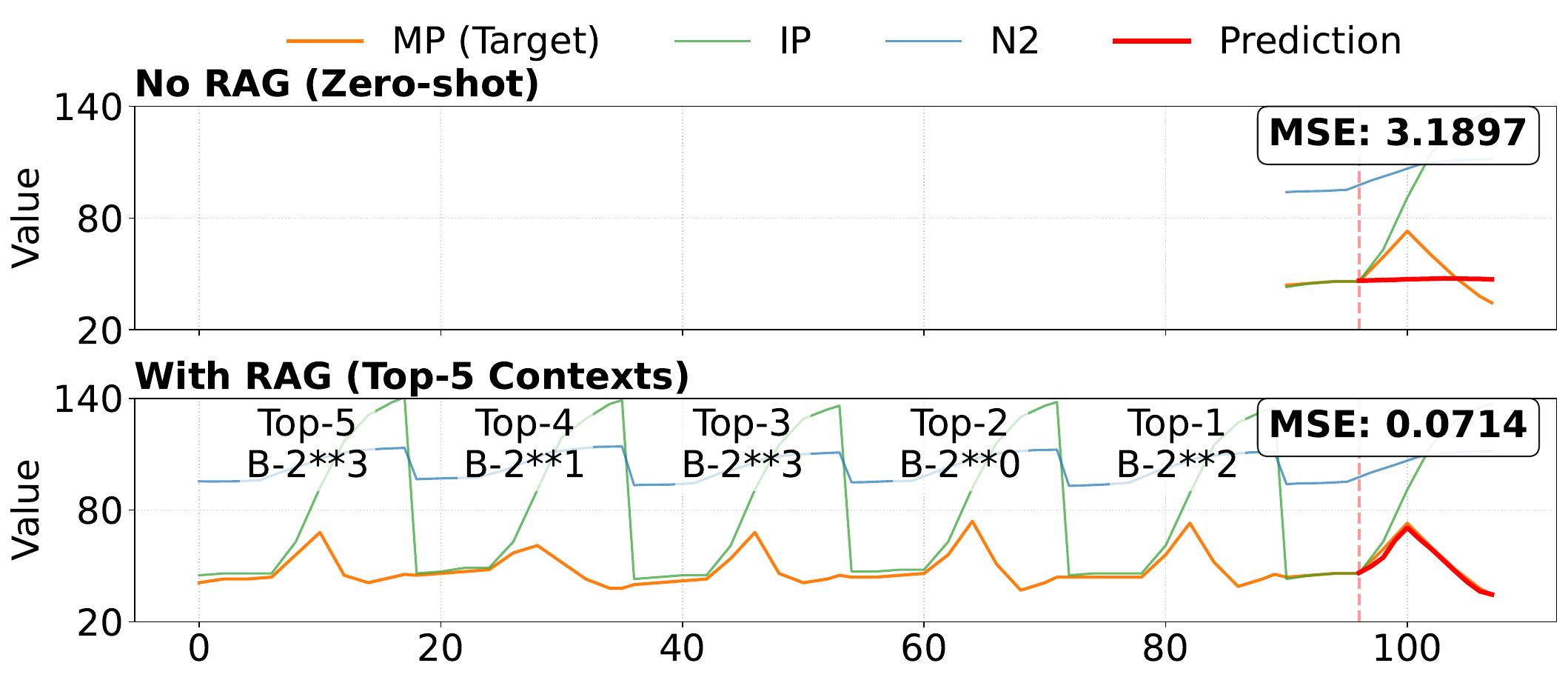}
    \caption{Visual comparison between the baseline and our RAG approach.}
    \label{fig:rag_vis}
\end{figure}

\begin{figure}
    \centering
    \includegraphics[width=\linewidth]{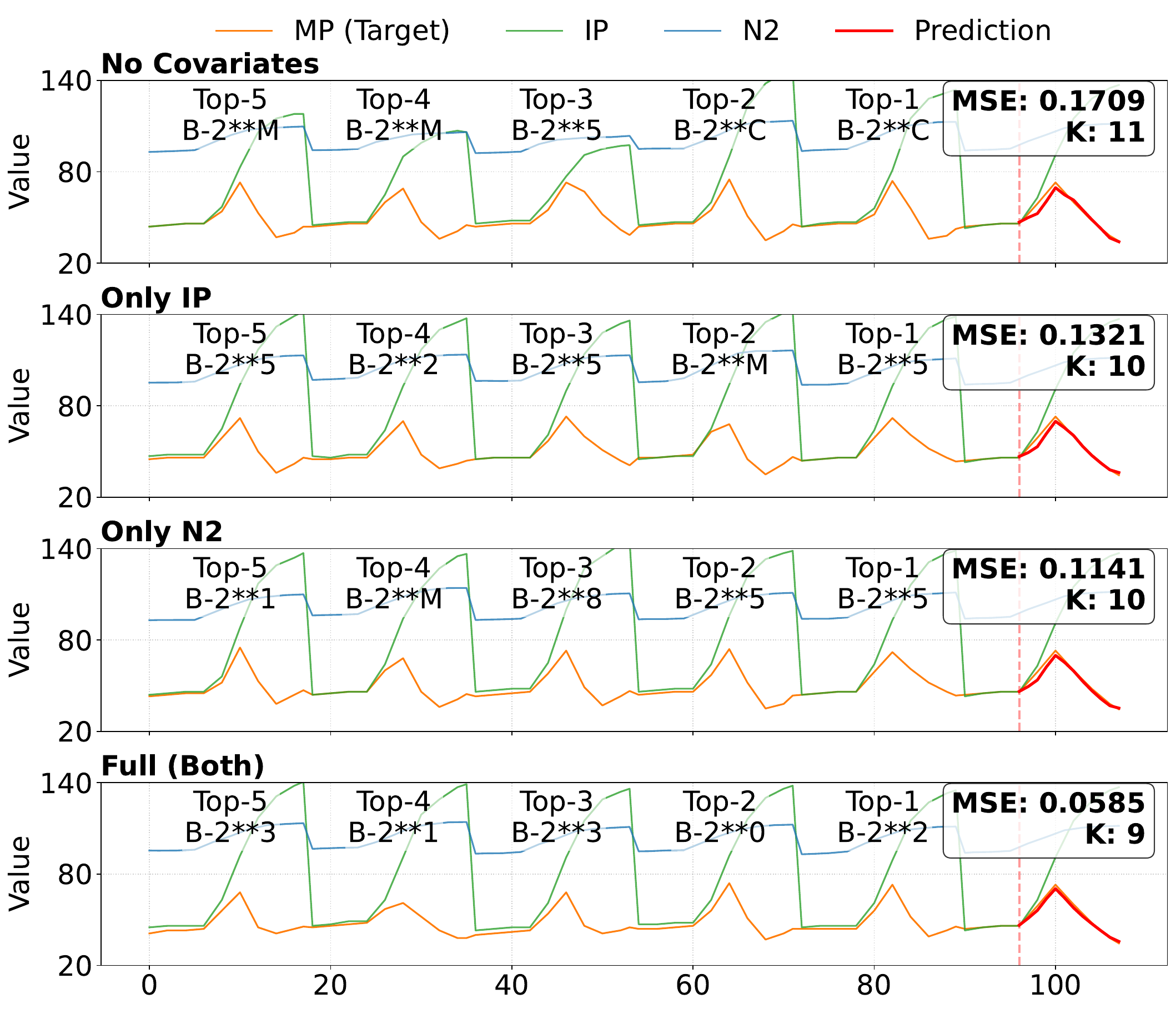}
    \caption{Visualization of covariate ablation studies. We compare prediction trajectories under four settings: (1) No Covariates, (2) Only IP, (3) Only N2, and (4) Full Covariates.}
    \label{fig:covariate_vis}
\end{figure}

\begin{figure}
    \centering
    \includegraphics[width=\linewidth]{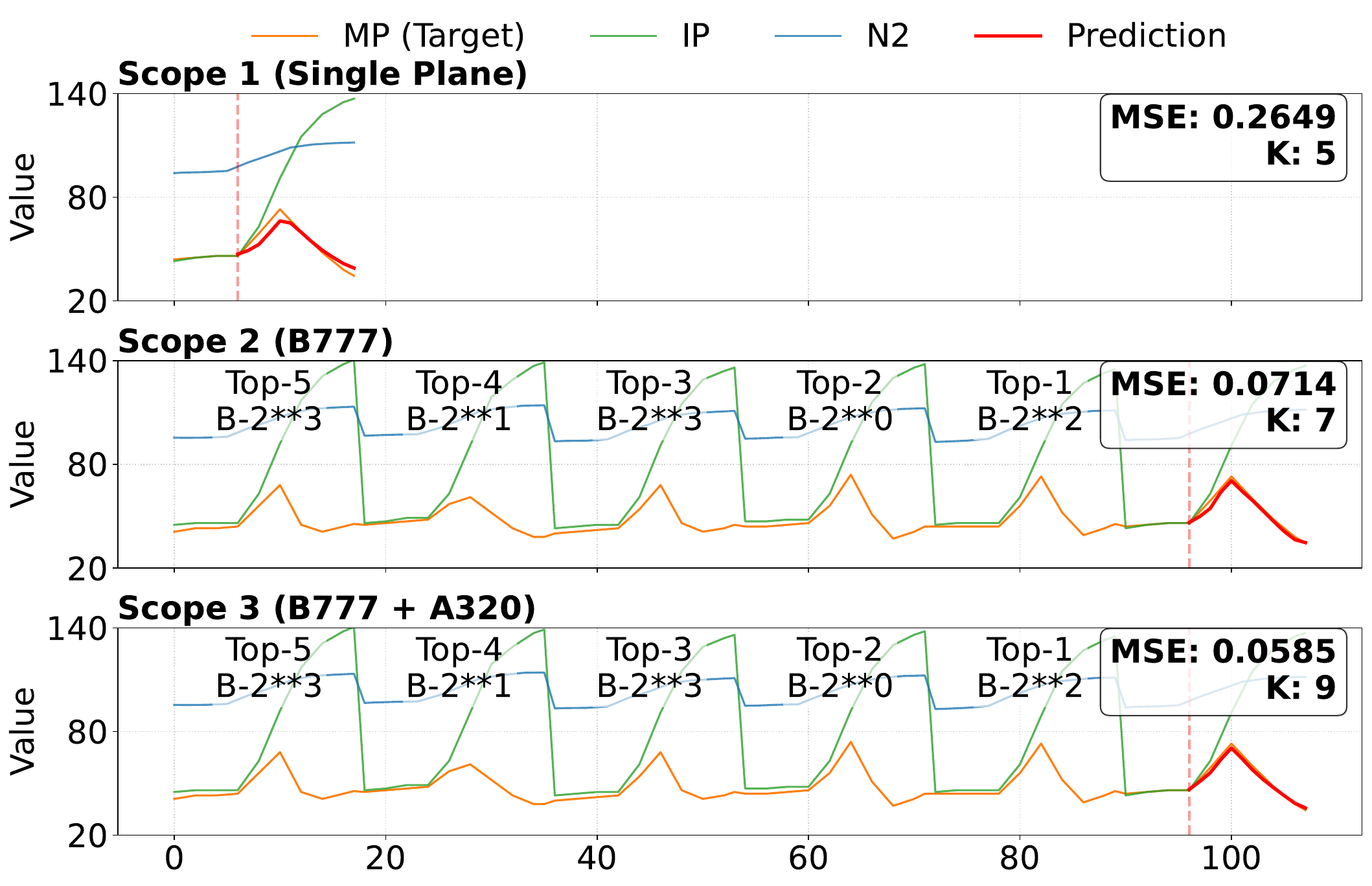}
    \caption{Impact of Knowledge Base (KB) Scope on retrieval and prediction. From top to bottom: Scope 1 (Single Plane), Scope 2 (B777, same aircraft type), and Scope 3 (B777 + A320, cross-Type mixed aircraft).}
    \label{fig:kb_scope_vis}
\end{figure}

\begin{figure}
    \centering
    \includegraphics[width=\linewidth]{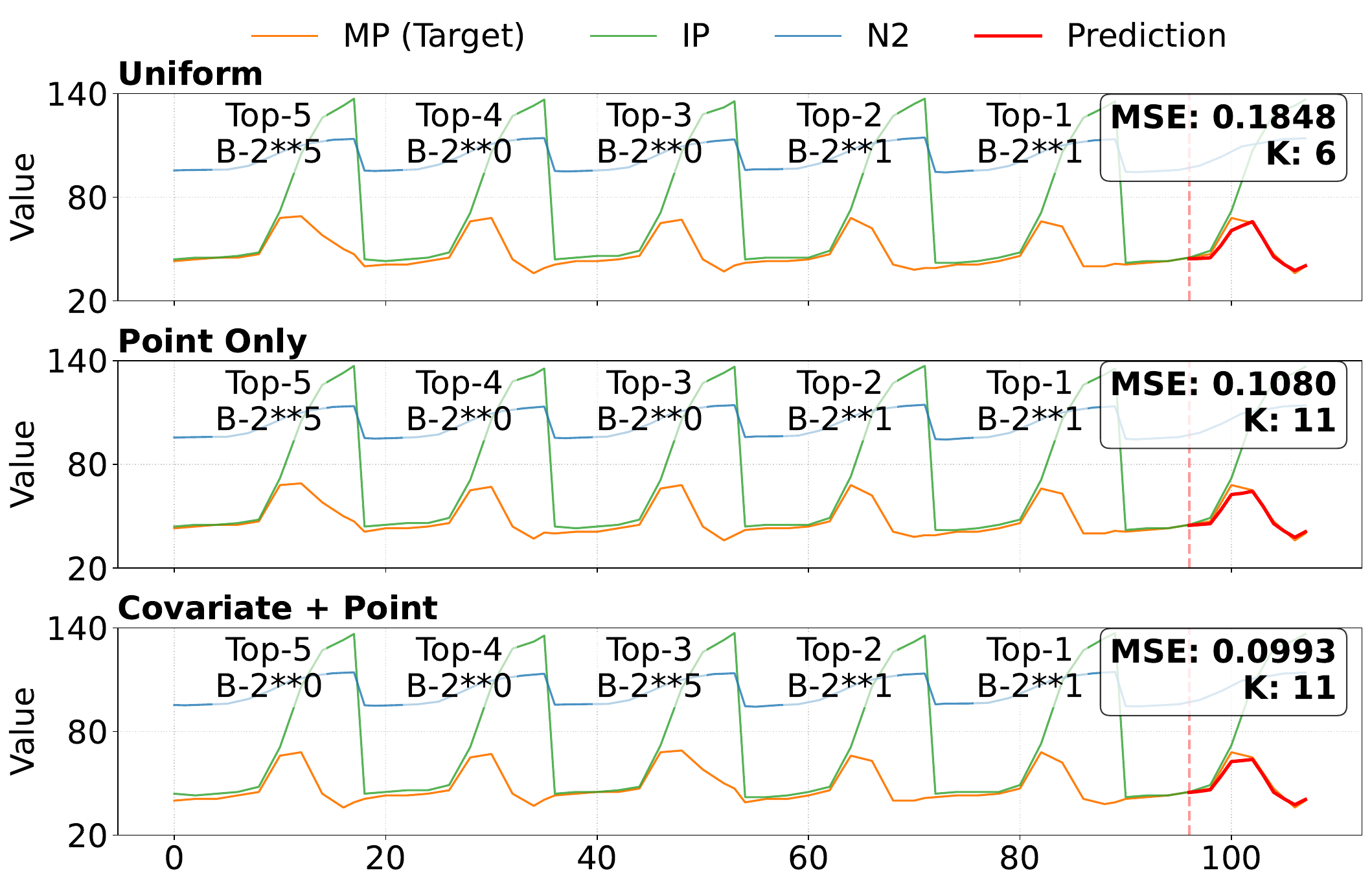}
    \caption{Comparison of weighting strategies: Uniform (Unweighted), Temporal Weight Only, and our proposed Variable + Temporal strategy.}
    \label{fig:weighting_vis}
\end{figure}

\begin{figure}
    \centering
    \includegraphics[width=\linewidth]{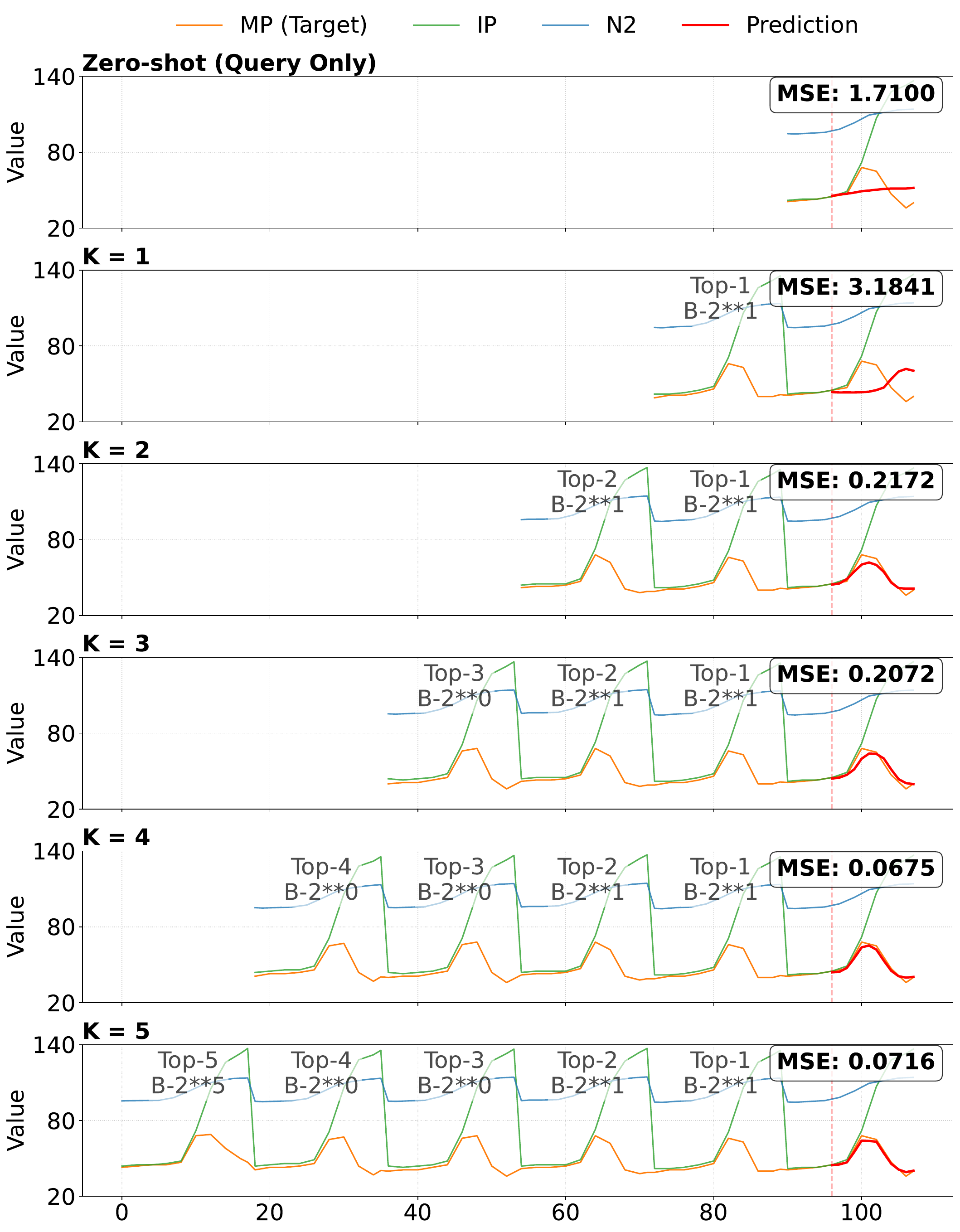}
    \caption{Visualization of prediction performance with varying context lengths (Top-$K$). The plots display the concatenated input sequences and resulting predictions for $K \in \{1, 2, 3, 4,5\}$.}
    \label{fig:k_vis}
\end{figure}

\begin{figure}
    \centering
    \includegraphics[width=\linewidth]{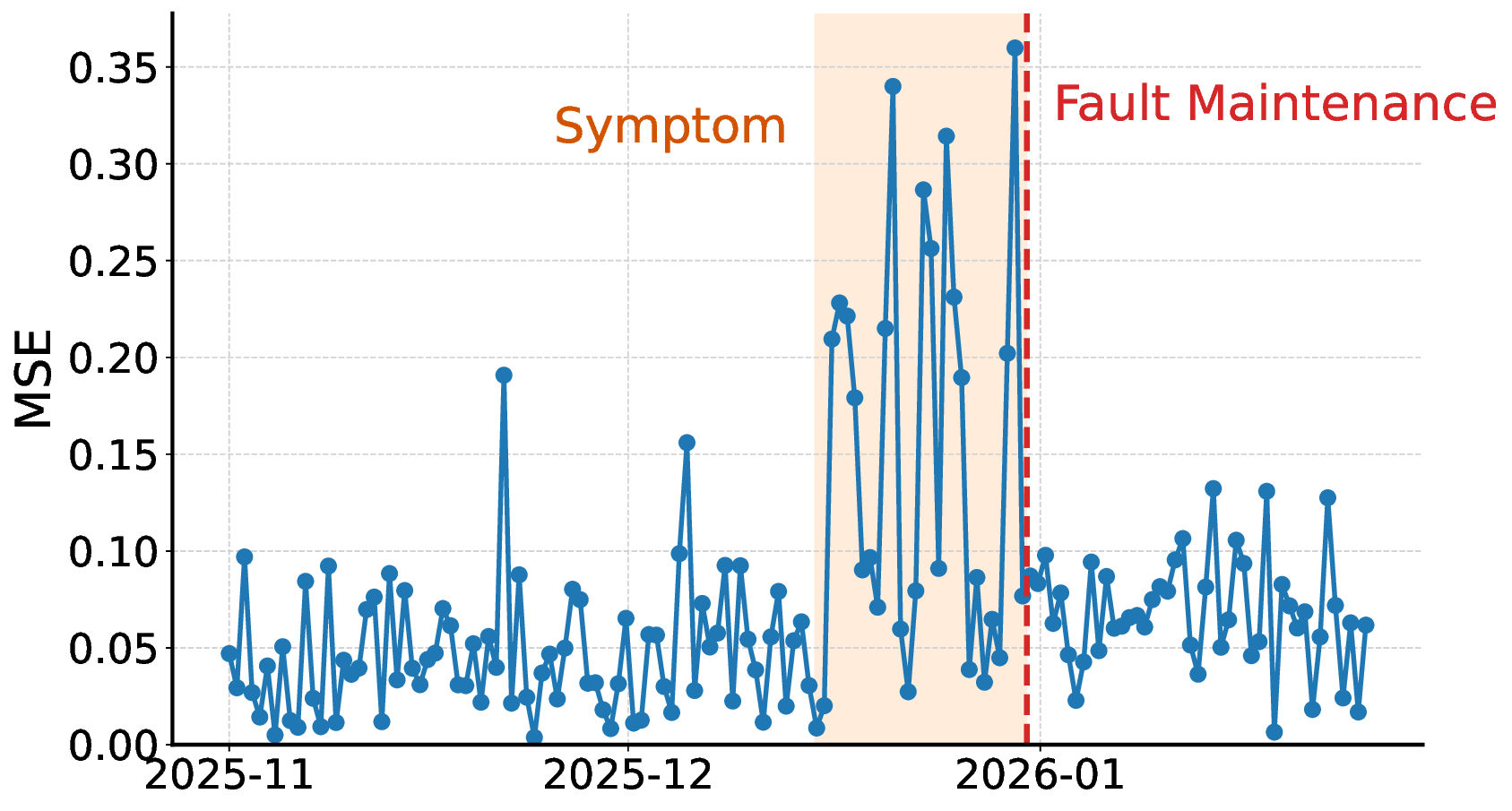}
    \caption{The MSE of the B-2**7 PRSOV L in time order. PRSOV L fault notified and detected with our method on plane B-2**7. }
    \label{fig:case_study_b2027}
\end{figure}

\subsection{RAG Visualization}
\label{appendix:rag_vis}
We visualize a specific query instance to demonstrate the efficacy of the Retrieval-Augmented Generation. 
Figure \ref{fig:rag_vis} compares the zero-shot baseline against our RAG model. 
While the baseline (Zero-shot) struggles to anticipate the sudden shift in physical parameters due to a lack of context, the RAG model successfully retrieves similar historical patterns (visualized as Top-5 contexts). 
The retrieved segments provide the necessary reference for the trend, significantly reducing the Mean Squared Error (MSE).

\subsection{Covariate Effect Visualization}
\label{appendix:covariate_vis}
Figure \ref{fig:covariate_vis} 
illustrates the impact of different covariate settings on trajectory prediction. 
We observe that the covariates contributes greatly to capture the complex inter-dependencies of the flight cycle. 
The ``Full'' setting (incorporating Maniford Pressure (target), Intermediate Pressure, and N2 speed) provides the necessary physical context, allowing the model to generate a trajectory that closely aligns with the ground truth, validating the importance of multi-variate modeling.

\subsection{Different KB Visualization}
\label{appendix:kb_vis}
We analyze how the scope of the Knowledge Base affects retrieval quality in Figure \ref{fig:kb_scope_vis}. 
Scope 1 (Single Plane)  offers high specificity but may suffer from data sparsity.
Scope 3 (B777) introduces distribution shifts that can degrade performance. 
Scope 2 (B777 + A320) strikes an optimal balance, providing sufficient diverse examples while maintaining physical consistency, as reflected in the lower MSE and better curve fitting.

\subsection{Weighted vs. Unweighted Visualization}
\label{appendix:weighting_vis}
Figure \ref{fig:weighting_vis} presents the ablation study on retrieval weighting strategies. 
By applying our proposed ``Covariate + Point'' weighting, the retriever prioritizes segments that are both temporally relevant and physically consistent with the query's current control inputs, filtering out noise and leading to more robust predictions.

\subsection{Different Top-k Visualization}
\label{appendix:topk_vis}

Figure \ref{fig:k_vis} visualizes the effect of context length ($K$) on the input sequence. 
We display the concatenated retrieved fragments preceding the query for different $K$ values. 
This visualization demonstrates the model's ability to attend to varying lengths of historical context augmentation. 
It highlights that a careful selection of concatenate segment is needed because an extra similar segment might not always be beneficial to the prediction.

\subsection{CSA PRSOV Maintenance Visualization}
\label{appendix:case_study_illustration}

Figure \ref{fig:case_study_b2027}
visualizes the Mean Squared Error (MSE) of our forecasts over time.
Since the RAG retriever is strictly constrained to a healthy Knowledge Base, the model acts as a physical consistency checker.
An inability to predict current MP status (High MSE) indicates a deviation from the ideal control logic.

Crucially, the highlighted orange regions reveal a cluster of \textbf{fault symptoms} rather than a constant failure.
In mechanical degradation, components often oscillate between normal operation and dysfunction.
Consequently, the MSE exhibits significant volatility, spiking during transient lapses in valve control and returning to baseline when the valve briefly functions correctly.
This distinct fluctuation pattern emerged decisively on December 31 for B-2**7, allowing maintenance teams to identify the degrading trend well before the ultimate functional fault.

\end{document}